\documentclass[manuscript]{acmart}

\usepackage{multirow}
\usepackage{subcaption}
\usepackage[noend]{algpseudocode}
\usepackage[nothing]{algorithm}
\renewcommand{\and}{\textbf{and} }
\renewcommand{\in}{\textbf{in} }
\newcommand{\is}{\textbf{is} }
\usepackage{amsmath}

\AtBeginDocument{%
  \providecommand\BibTeX{{%
    \normalfont B\kern-0.5em{\scshape i\kern-0.25em b}\kern-0.8em\TeX}}}

\setcopyright{acmcopyright}
\copyrightyear{2018}
\acmYear{2018}
\acmDOI{10.1145/1122445.1122456}

\acmConference[FDG'20]{FDG'20: Foundations of Digital Games}{September 15--18, 2020}{Malta}
\acmBooktitle{FDG'20: Foundations of Digital Games,
 September 15--18, 2020, Malta}
\acmPrice{15.00}
\acmISBN{978-1-4503-XXXX-X/18/06}

\copyrightyear{2020}
\acmYear{2020}
\setcopyright{acmcopyright}\acmConference[FDG '20]{International
Conference on the Foundations of Digital Games}{September 15--18,
2020}{Bugibba, Malta}
\acmBooktitle{International Conference on the Foundations of Digital
Games (FDG '20), September 15--18, 2020, Bugibba, Malta}
\acmPrice{15.00}
\acmDOI{10.1145/3402942.3402955}
\acmISBN{978-1-4503-8807-8/20/09}

\begin{document}

\title{Automatic Critical Mechanic Discovery \\ Using Playtraces in Video Games}

\author{Michael Cerny Green}
\affiliation{%
  \institution{New York University; OriGen.AI}
  \streetaddress{370 Jay Street}
  \city{Brooklyn}
  \country{USA}
}
\email{mike.green@nyu.edu}

\author{Ahmed Khalifa}
\affiliation{%
  \institution{New York University}
  \streetaddress{370 Jay Street}
  \city{Brooklyn}
  \country{USA}
}
\email{ahmed@akhalifa.org}

\author{Gabriella A. B. Barros}
\affiliation{%
  \institution{Modl.AI}
  \city{}
  \country{Brazil}
}
\email{gabbbarros@gmail.com}

\author{Tiago Machado}
\affiliation{%
  \institution{Northeastern University}
  \city{Boston}
  \country{USA}
}
\email{tiago.machado@nyu.edu}

\author{Julian Togelius}
\affiliation{%
  \institution{New York University}
  \streetaddress{370 Jay Street}
  \city{Brooklyn}
  \country{USA}
}
\email{julian@togelius.com}

\renewcommand{\shortauthors}{Green et al.}

\begin{abstract}
  We present a new method of automatic critical mechanic discovery for video games using a combination of game description parsing and playtrace information. This method is applied to several games within the General Video Game Artificial Intelligence (GVG-AI) framework. In a user study, human-identified mechanics are compared against system-identified critical mechanics to verify alignment between humans and the system. The results of the study demonstrate that the new method is able to match humans with higher consistency than baseline. Our system is further validated by comparing MCTS agents augmented with critical mechanics and vanilla MCTS agents on $4$ games from GVG-AI. Our new playtrace method shows a significant performance improvement over the baseline for all $4$ tested games. The proposed method also shows either matched or improved performance over the old method, demonstrating that playtrace information is responsible for more complete critical mechanic discovery.
\end{abstract}



\maketitle

\section{Introduction}

Tutorials are designed to help a player learn how to play a game. They come in several different forms, such as text instructions (e.g. ``press A to jump''), examples where an agent demonstrates what to do (e.g. watching an AI jump), and interactive content, like levels, that gradually introduce game mechanics as you play them. They are often the player's first contact with the game, and a player's experience with a tutorial can strongly impact their opinion of said game. 

The ability to automatically or semi-automatically generate tutorials would be significant to developers, as most tutorials are made manually. Outside of the time/cost savings a system like this would allow, automated tutorial generation would expand upon the potential for fully automatic game generation, as previous attempts so far have demonstrated that evaluating generated games for humans, without using human-like playing ability~\cite{nielsen2015towards,cook2014ludus} is not trivial. However, in order to generate a game tutorial, a system would first need to identify what content should be taught. Automatically finding important mechanics may provide insight into game design itself, showing developers new ways of playing a game or of measuring game qualities, such as the game's depth~\cite{lantz2017depth}.

Games tend to utilize a combination of tutorial styles to teach important features. Previous research in automatic tutorial generation has defined possible tutorial types~\cite{green2017press} and methods for generating tutorial text~\cite{green2018atdelfi}, visual demonstrations~\cite{green2018atdelfi}, and levels~\cite{green2018generating,khalifa2019intentional}. The \emph{AtDelfi} system
\footnote{https://github.com/mcgreentn/GVGAI} 
uses search methods to automatically identify the critical mechanics of a game~\cite{green2018atdelfi}. We define ``critical mechanics'' as the set of mechanics necessary to trigger in order to win a level. 
In other words, every winning playthrough will contain this set of mechanics\footnote{One could imagine a scenario where the player could have multiple choices in a level, resulting in a disjointed set of critical mechanics, depending on the gameplay path selected.}. 
The mechanic discovery method in \emph{AtDelfi} was simple and somewhat successful, but had shortcomings. In this paper, we propose an improved method for automatically identifying critical mechanics in games. 

A complicated task, such as playing a video game, can often be divided into a number of subtasks, each with their own subgoals. For example, leaving a room might involve finding a key, removing any obstacles on the way to the door, getting to the door and opening it. The idea of subdividing a larger task into smaller constituent tasks in order to make it easier to solve is common within both the planning and reinforcement learning literature~\cite{mcgovern2001automatic,asadi2005autonomous}. One can find similar ideas in the work presented here, where subgoals are restricted to the triggering of specific game mechanics, rather than finding individual game states.

In this paper, we demonstrate a new method for the automatic discovery of ``critical game mechanics'' using playtraces from humans and/or artificial agents, 
and recommend this as a module within a tutorial generator system. We evaluate this approach through a two-step process. First, we present an user study that compares which mechanics humans believe to be critical against the \emph{AtDelfi} method and the new method. Secondly, we demonstrate a new way of incorporating mechanic information into stochastic forward planning algorithms, such as Monte Carlo Tree Search~\cite{browne2012survey}, which we use to compare a baseline MCTS agent and agents with mechanic information taken from each discovery method.





\section{Background}
The following section discusses previous research in the areas of Monte Carlo Tree Search (MCTS) 
automated tutorial generation and critical mechanic discovery, 
subgoal discovery in reinforcement learning and hierarchical planning, and the General Video Game AI framework.



\subsection{Monte Carlo Tree Search (MCTS)}
MCTS~\cite{browne2012survey,coulom2006efficient,kocsis2006bandit} is a stochastic tree search algorithm that creates asymmetric trees by expanding more promising nodes more often. It consists of four phases: \textit{selection}, \textit{expansion}, \textit{simulation}, and \textit{backpropagation}. In the \emph{selection phase}, the algorithm decides which node it should select to expand next using a \textit{selection policy}, a popular choice being UCB1~\cite{kocsis2006improved}. This policy defines how the algorithm will \textit{select} between exploring or exploiting nodes. During the \emph{expansion phase}, a new node is added to the tree as a child of the selected node. During the \emph{simulation phase}, the newly created child node is forward-simulated until it reaches either some terminal state (a win or a loss) or some pre-defined threshold (i.e 500 moves into the future). Finally, in the \emph{backpropagation phase}, the reward value is calculated for the simulation phase's final state and is used to update the values of the visited nodes, from the newly created node to the tree root. The algorithm runs in an iterative fashion, and the updated node values define how to guide the search in the next iteration.

MCTS can be improved depending on the environment. Macro actions~\cite{perez2014solving} and mixmax~\cite{jacobsen2014monte} are some examples. UCT functions can even be evolved for general~\cite{bravi16evolving} or specific~\cite{holmgard2018automated} environments/playstyles. Inspired by this, the agents in this paper contained modified reward equations.


\subsection{Tutorial Generation and Critical Mechanic Discovery}
Several projects have addressed challenges in automatic tutorial generation, such as heuristic generation for Blackjack and Poker \cite{de2016generating,de2018flop,de2018texas} or quest/achievement generation in \emph{Minecraft}~\cite{alexander2017deriving}. Mechanic Miner~\cite{cook2013mechanic} is able to evolve simple mechanics for 2D puzzle-platform games using \emph{Reflection}\footnote{https://code.google.com/archive/p/reflections/}, which it uses to generate levels. The \emph{Gemini} system~\cite{summerville2017mechanics} takes game mechanics as input and performs static reasoning to find higher-level meanings about the game. Similarly, Mappy~\cite{osborn2017automatic} receives a Nintendo Entertainment System game and a series of button presses as input, and generates a graph of room associations, transforming movement mechanics into information. 

\begin{figure}
    \centering
    \includegraphics[width=\columnwidth]{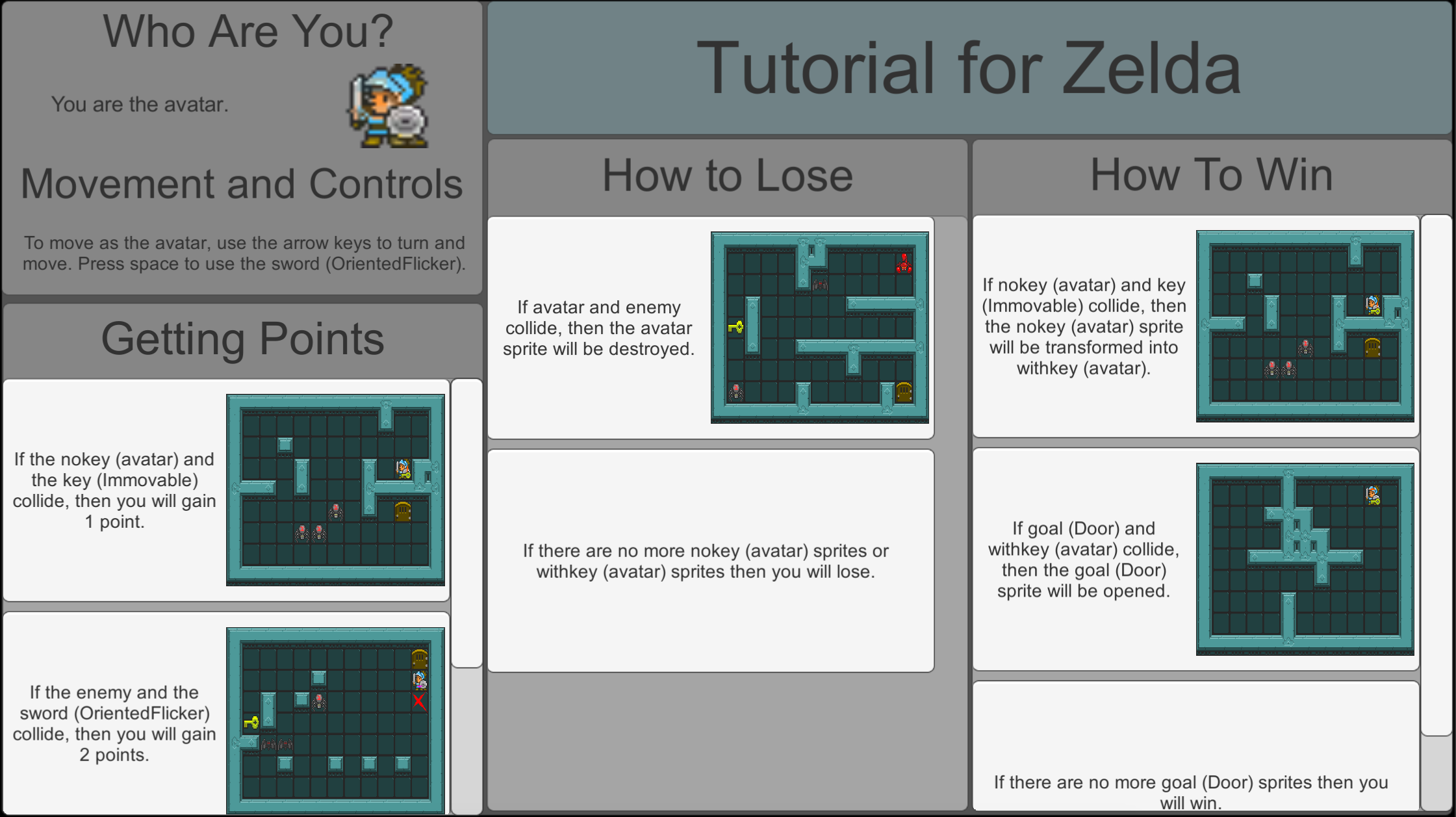}
    \caption{An AtDelfi generated tutorial for GVGAI's Zelda}
    \label{fig:atdelfi}
\end{figure}

The \emph{AtDelfi} system~\cite{green2018atdelfi} attempts to solve the challenge of automatically finding critical mechanics for the purpose of generating tutorials. Figure~\ref{fig:atdelfi} displays a tutorial card generated by the system. In addition to finding critical mechanics, \emph{AtDelfi} also includes mechanics that reward points or lead to the player losing the game. Generated tutorials explain selected mechanics through text and GIFs. We describe in detail how \emph{AtDelfi} finds critical mechanics at the end of Section \ref{sec:overview-search}.




\subsection{Subgoals in Reinforcement Learning and Planning}\label{sec:subgoals}
Singh~\cite{singh1992transfer} proposed the existence of \emph{elemental tasks}, i.e. behaviors an agent can achieve that accomplish some conditional goal. By sequentially lining up these elemental tasks, an agent could improve training and generalization by using what it learned to overcome the previous task to tackle the next.

Subgoal discovery builds on the idea of a state or behavior marking progress along the path of solving a problem. The goal is to automatically derive intermediate reward \textit{states} to improve performance. Maron's Diverse Density Algorithm~\cite{maron1998framework} 
was first used for automated subgoal discovery by McGovern and Barto~\cite{mcgovern2001automatic}. Asadi and Huber used Monte Carlo sampling in reinforcement agents to discover subgoals for faster training~\cite{asadi2005autonomous}.

Hierarchical MCTS algorithms~\cite{vien2015hierarchical} typically take advantage of information gathering to automatically find target states to assist in the building of the search trees of agents, such as UCT and partially observable Markov decision process (POMDP) agents. This approach works particularly well when a Markov decision process is abstracted into a partially observable one, as this can significantly reduce the state branching factor~\cite{bai2016markovian}. IGRES is an example of a randomized POMDP solver that uses subgoal discovery to leverage information about state space~\cite{ma2015information}. IGRES is able to cut down on potential solution space, thus decreasing the amount of computation time while maintaining good performance. 

It is important to note that the method proposed in this paper is not intended as a contribution to hierarchical planning; rather the MCTS experiment within is carried out as a way of evaluating a critical mechanic discovery method.

\subsection{General Video Game Artificial Intelligence Framework (GVG-AI)}\label{sec:background-GVGAI}

GVG-AI is a framework for general video game playing~\cite{perez2016general,perez2019general}, aimed at exploring the problem of creating artificial players that are able to play a variety of game. It has an annual competition where AI agents take part and are judged on their performance in games unseen by them beforehand. In the competition, each agent has to decide the next taken action in 40 milliseconds provided with a forward model for the current game. 
The framework's environment is constantly evolving~\cite{perez2019general} and adding more tracks to the competition, such as level generation track~\cite{khalifa2016general}, rule generation track~\cite{khalifa2017general}, learning agents track~\cite{torrado2018deep}, and two-player agents
track~\cite{gaina2016general}.

The GVG-AI framework uses the Video Game Description Language (VGDL) to describe the games it runs~\cite{ebner2013towards}. The language is human-readable, simple and compact, but expressive enough to allow for the creation of a wide variety of simple 2D games. Some of them are adaptations of classical games, 
such as \emph{Pacman} (Namco 1980) and \emph{Sokoban} (Imabayashi 1981), while others are brand new games, such as \emph{Wait For Breakfast}. To write a game in VGDL, one only needs to describe the behaviour of game elements, what happens when they collide, and how to win or lose the game. A VGDL game consists of a game description file and one or more level description files. The game description file contains a Sprite Set, or game objects that can be instantiated, including the sprite's behavior, images used, etc; an Interaction Set, or a list of how sprites interact; a Termination Set, or what conditions trigger an end to the game; and a mapping between game sprites and the symbols representing them in the level files.

\section{System Overview} 
Our system receives two inputs: a game description file that contains the game rules in VGDL and a series of playtraces of the game. Using the game description, it builds a ``mechanic graph'', which contains the system's understanding of all game rules. It inserts playtrace data into this graph, then searches it to find ``critical mechanics.'' A ``mechanic'' can be defined as an event within the game that is fired by a game element that impacts the game's state~\cite{sicart2008defining}.
For this work, we assume that there is a single linear path the player must follow through the level. ``Critical mechanics'' are the mechanics necessary to trigger in order to win a level. We then augment MCTS agents with these mechanics by modifying their state evaluation function to take into account the occurrence of these mechanics during play. The following subsections further describe the mechanic graph creation, the playtrace informed graph search, and the modifying of an MCTS agent with mechanic information.

\subsection{Mechanic Graph Generation}\label{sec:overview-graph_gen}
\begin{figure}
    \centering
    \includegraphics[width=0.9\linewidth]{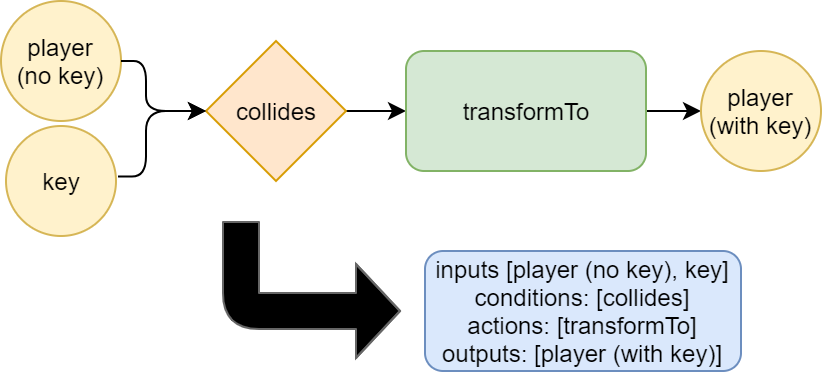}
    \caption{A collection of nodes representing a pickup-key mechanic. A player colliding with a key results in the player picking up the key. This can be transformed into a single \emph{mechanic node}.}
    \label{fig:example-mechanic}
\end{figure}

\begin{figure}
    \centering
    \includegraphics[width=0.9\linewidth]{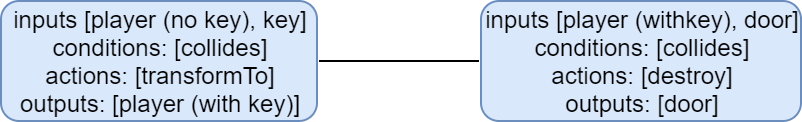}
    \caption{An example of how mechanic nodes that share inputs/outputs are linked using an edge. The shared I/O is player (withkey).}
    \label{fig:example-mechanic-linkage}
\end{figure}

The first step of critical path construction involves the mechanics of the game in question. Our system contains the same parser as the one in the \emph{AtDelfi} system~\cite{green2018atdelfi}, which is able to transform VGDL code into an ``atomic interaction graph,'' which contains game objects (e.g. sprites and other objects), conditions (e.g. collisions, termination, etc), and events that occur if these conditions are met (e.g. destroying a sprite, gaining points, etc). Please note that the atomic interaction graph was known as a ``mechanic graph'' in the original \emph{AtDelfi} paper~\cite{green2018atdelfi}; we have selected to rename it in reference to better articulate its purpose. All internal types of objects, conditions, and action nodes in the atomic interaction graph are derived directly from VGDL language. Figure \ref{fig:example-mechanic} displays an example of a player picking up a key as seen by the system after parsing VGDL for building an atomic interaction graph. The system then abstracts these node elements into a ``mechanic graph,'' where each mechanic is represented as a single node. This abstraction is done to better organize the search space into concretely defined mechanics nodes, in contrast to the atomic interaction representation. In a mechanic graph, any object, condition, or action can be a part of a mechanic node, but they do not exclusively belong to a mechanic.  For example, a player object can be a member of a "pickup key" mechanic node, as well as an "open door" mechanic node. To complete this transformation, the algorithm loops over all nodes in the atomic interaction graph; object nodes that are linked directly to a unique condition-action node pair are considered a single mechanic. Mechanics which share input and/or output game objects are linked using an edge, see Figure \ref{fig:example-mechanic-linkage}.


\subsection{Critical Mechanic Search}\label{sec:overview-search}
In this paper, we compare two different methods of critical mechanic discovery. One is the new playtrace method we present in this paper; the second is the method used in the original \emph{AtDelfi} system.
\subsubsection{Playtrace Method}
After a mechanic graph is created and all possible game mechanics are represented, the system informs the graph with playtrace information, which can be collected from human players or automated agents. Given a collection of playtraces for a single game level, the system looks for the playtrace that (1) contains the lowest amount of unique mechanics represented on the graph, \emph{and} (2) in which the player won the level. In doing so, it infers that the playtrace must contain knowledge of which mechanics must be triggered in order to beat the level. By singling out the playtrace with the lowest amount of unique mechanics, it can minimize gameplay ``noise'', such as accidentally walking into walls (which triggers an interaction with the wall), or triggering other events that have nothing to do with winning the game. 

Each mechanic in the playtrace is linked to the particular game-frame in which it occurred. For each unique mechanic triggered during that playtrace, the system looks for the earliest frame during gameplay when that mechanic occurred and enters it into the corresponding node in the graph. Once this has been done for all mechanics, the system performs a modified best first search algorithm over the graph, starting from player-centric mechanics (i.e. those that the player either initiates or is otherwise involved in, like colliding with coins or swinging a sword) and ending with a positive terminating one (i.e. winning the game). The algorithm thus behaves like a greedy best first algorithm that records all mechanics visited until reaching a terminal one. The cost of a node is that node's frame value. The algorithms ends if the current picked node is a terminal node. The pseudocode for this process can be found in Algorithm~\ref{fig:psuedocode}.

\begin{algorithm}
\caption{Finding the Critical Path of a Game}\label{fig:psuedocode}
\begin{algorithmic}[1]
\Function{findCritPath}{}
\State $\textit{searchlist} \gets \text{getAllPlayerCentricMechanics()}$
\State $criticalPath \gets \textit{[]}$
\While{$searchList != \textit{[]}$}
\State sortAscending($searchList$, frame) 
\State $current \gets \textit{searchList[0]}$
\State $searchList.remove(current)$
\State $criticalPath.add(current)$
\If{$current$ \is $WIN$}
\State $break$
\EndIf
\For{$n$ \in $current.neighbors()$}
\If{$n$ != $VISITED$ \& $n.frame$ >= $current.frame$}
\State $searchList.add(n)$
\EndIf
\EndFor
\EndWhile
\State \Return{$criticalPath$}
\EndFunction
\end{algorithmic}

\end{algorithm}

Thus, the search creates a path of the earliest occurring mechanics, which then becomes the list of the game's critical mechanics. Additionally, the system also automatically adds any ``sibling-mechanics,'' or mechanics that are nearly identical in nature to ones in the critical mechanic list, to the list. Sibling mechanics are mechanics that contain identical condition-action pairs, and sprites that are classified in VGDL having the same parent~\footnote{https://github.com/GAIGResearch/GVGAI/wiki/Sprites}. For example, in GVG-AI's Zelda, hitting either a bat or a spider with the sword results in that entity's destruction. In Zelda description file, bat and spiders are both identified under a single parent (enemy). If either the bat-sword mechanic or the spider-sword mechanic is contained within the critical mechanic list, the other one will also be included. 

\subsubsection{\emph{AtDelfi} Method}
As contrast, the ``\emph{AtDelfi}-method'' referenced in this paper refers to the method of critical mechanic discovery in the current iteration of the \emph{AtDelfi} system~\cite{green2018atdelfi}. This method only uses the game description file as input in order to generate an atomic interaction graph, as described before, and does not inject playtrace information into this graph. To find critical mechanics, the system searches through the interaction graph using a simple Breadth First Search algorithm, looking for the shortest path between a player-driven condition node to the winning terminal action node. The longest of these ``shortest paths'' would be selected as the critical path, and the interaction nodes are then transformed into mechanics using the method described in Section \ref{sec:overview-graph_gen} and displayed in Figure \ref{fig:example-mechanic}. 

\subsection{Mechanic-Augmented MCTS}\label{sec:overview-mechanic}
After the critical mechanics for a particular game have been found (either using the play traces method or AtDelphi method), we can augment an MCTS agent with this mechanic information. Traditionally, the evaluation function of an MCTS agent takes into account the game state at the end of the \emph{simulation} phase of the algorithm, and then the reward is backpropagated up the tree. However, we can modify this evaluation function to take into account all simulated event data as well, adding additional rewards for any simulated events that match conditions of critical path mechanics. This is a similar approach to the use of subgoals in hierarchical planning~\cite{vien2015hierarchical} mentioned previously in the background section, one difference being that the agent is a simple MCTS agent rather than a more complex hierarchical MCTS or reinforcement learning agent. Another notable difference is that the subgoals defined here are represented as game mechanics, rather than game states. This partial state abstraction affords a greater degree of generality across domains.

The value of these additional mechanic rewards decreases with frequency. Each time the agent triggers a specific mechanic in its past during play, the subsequent reward decreases by $1/frequency$, in order to both encourage the agent to trigger multiple mechanics, and to discourage the agent to keep triggering the same mechanic repeatedly. This reward is also decreased the further out in planning the agent finds the mechanic, similar to discount factors in reinforcement learning. Therefore, mechanics triggered earlier on in planning backpropagate greater rewards than those that happened later. This allows an agent to better focus its search to areas where mechanics trigger early and frequently. The reward equation for a single instance of a critical mechanic during planning is given in Equation~\ref{eq:reward}, where \textit{F} is the number of occurrences this mechanic has been triggered until now, \textit{$T_{current}$} is the in game frame where the mechanic was triggered, and \textit{$T_{root}$} is game frame at the root node.

\begin{equation}
    R = \frac{1}{F * 1.1^{T_{current} - T_{root}}}
    \label{eq:reward}
\end{equation}

\section{Experiments}\label{sec:experiments}

This system is designed to accept both human and agent playtraces, as long as the game can be beaten. Thus, we collected human playtraces to run experiments on the algorithm for creating critical paths of mechanics. Participants played a minimum of $3$ different levels each for $4$ GVGAI games: 
\begin{itemize}
    \item \textbf{Solarfox:} is a port of \emph{Solar Fox} (Bally/Midway Mfg. Co 1981). The goal is to collect all the gems in the level, while dodging the flames being thrown by enemies. Each gem collected gives the player a point. Several levels contain ``powered gems,'' which are worth no points. If a player collides with a powered gem, it will spawn a ``gem generator,'' which can generate more gems to collect and gain more points. If a player touches a generator, however, the generator will be destroyed and no longer generate any more gems. 
    \item \textbf{Zelda:} is inspired by \emph{The Legend of Zelda} (Nintendo 1986). To win, the player must pick up the key and unlock the door. Monsters populate the level and can kill the player, causing them to lose. The player can swing a sword; if the sword hits a monster, the monster is destroyed, and the player gains a point.
    \item \textbf{Plants:} is inspired by \emph{Plants vs. Zombies} (PopCap Games 2009). If the player survives for $1000$ game ticks, they win. Zombies spawn on the right side of the screen and move left. The player loses if a zombie reaches the left side. The player needs to grow plants on the left side of the screen. Plants automatically fire zombie-killing peas. Each zombie killed is worth a point. Occasionally, zombies will throw axes, which destroy plants.
    \item \textbf{RealPortals:} is inspired by \emph{Portal} (Valve 2007). The player must reach the goal, which sometimes is behind a locked door that needs a key. Movement is restricted by water, which kills the player if they touch it. To succeed, players need to pick up wands, which allow them to toggle between the ability to create \emph{portal entrances} and \emph{portal exits} through which they can travel across the map. There are also potions on some levels, which the player can push into the water to transform the water into solid ground.
\end{itemize}

These games were selected based on previous work~\cite{bontrager2016matching}, which categorized these games as ones that MCTS algorithms perform particularly poorly on. They also contain a diverse array of mechanics, terminal conditions (time-based (Plants), lock-and-key (Zelda and RealPortals), and collection (SolarFox)), and ranging levels of complexity.

The system runs with four games an average of 23 human playtraces for each game. In Table~\ref{tab:userstudyResults}, the ``Playtrace Method'' and ``AtDelfi Method'' columns show the identified critical mechanics for each of these games marked as ``X''s. This table was made using raw mechanic information output by our system, translated by humans into a more understandable form. For example, the original game rule ``door avatar(withkey) KillSprite'' essentially means ``Unlock the door with a key.'' The system attempts to find the minimum number of mechanics that are important in order to win. For example: the discovered critical mechanics for Zelda do not include any related to destroying enemies, because the player does not need to destroy enemies to win (unless they are blocking their way). 

\section{Evaluation}\label{sec:evaluation}
Before a critical path of mechanics could be used by another system (such as for the creation of tutorials), it is necessary to verify if the subgoals/mechanics in the path are actually ``critical,'' i.e. are important in order to achieve a good performance in the game. We propose a two-step evaluation method to do this for critical mechanic discovery methods.

First, a user study compares human-identified critical mechanics against the system-identified ones. The user study experiment evaluates how closely a method matches what humans identify as critical mechanics.  Second, identified critical mechanics can be inserted into MCTS reward functions. The agent-comparison experiment verifies that (at least from the perspective of a game-playing artificial agent) triggering critical mechanics discovered by a method results in better agent performance. The following subsections explain the human-identified mechanic comparison study and present the results of MCTS agent comparison study in detail.

\subsection{Human-identified Mechanic Comparison Study}\label{sec:validation-userstudy}
In the user study, we compare system-discovered critical mechanics to human-identified ones. The study participants were chosen by sending out a university-wide email to students asking for participation, as well as forwarding to friends and colleagues at other universities. Demographic information about the $93$ participants is shown in Table \ref{tab:demographics}. We compare the method proposed in this paper to the one used in the \emph{AtDelfi} system~\cite{green2018atdelfi} as a baseline.

\begin{table}
\centering
\begin{tabular}{|ccc|cccc|}
\hline
\multicolumn{3}{|c|}{\textbf{Age}} & \multicolumn{4}{c|}{\textbf{Game Playing Frequency}} \\ 
\textless{}25  & 25-34   & 35+    & None    & Casually  & Often   & Everyday  \\ \hline
24.7\%         & 68.8\%  & 6.5\%  & 4.3\%   & 36.6\%    & 16.1\%  & 43.0\% 
\\ \hline
\end{tabular}
\caption{User study participant demographics}
\label{tab:demographics}
\end{table}

Our user study application displayed a prompt describing the study's purpose. After completing the levels of a game, participants would be given the following prompt: ``In short sentences, describe what the player needs to do in order to perform well in the game.'' The participants responded using a free-text answer space. We deliberately chose the prompt wording and the answer space to avoid biasing the players, which might have happened if we had explicitly defined a mechanic or a critical mechanic.

\begin{table*}[ht]
    \centering
    \resizebox{1\textwidth}{!}{%

\begin{tabular}{|c|l|c|c|c|}
\hline
\textbf{\textbf{Game}} & \textbf{Mechanic}                                                 & \textbf{Percentage} & \multicolumn{1}{l|}{\textbf{Playtrace Method}} & \multicolumn{1}{l|}{\textbf{Baseline Method}} \\ \hline
\multirow{4}{*}{Solarfox}               & Avoid Flames                                                                       & 68\%                                 &                                                                          &                                                                \\
                                        & Collide with gems to pick them up                                                  & 64\%                                 & X                                                                        & X                                                              \\
                                        & Avoid Walls                                                                        & 18\%                                 &                                                                          &                                                                \\ \cline{2-5} 
                                        & \textbf{Match Rate}                                                                         & \textbf{-}                                    & \textbf{45.45}\%                                                                  & \textbf{45.45}\%                                                        \\ \hline
\multirow{7}{*}{Zelda}                  & Collide with the key to pick it up                                                 & 80\%                                 & X                                                                        & X                                                              \\
                                        & Unlock the door with the key                                                       & 80\%                                 & X                                                                        & X                                                              \\
                                        & Kill Enemies with Sword                                                            & 76\%                                 &                                                                          &                                                                \\
                                        & Avoid dying by colliding with Enemies                                              & 60\%                                 &                                                                          &                                                                \\
                                        & Navigate the level walls using arrow keys                                          & 20\%                                 &                                                                          &                                                                \\
                                        & Move quickly                                                                       & 12\%                                 &                                                                          &                                                                \\ \cline{2-5} 
                                        & \textbf{Match Rate}                                                                & \textbf{-}                           & \textbf{48.8\%}                                                          & \textbf{48.8\%}                                                \\ \hline
\multirow{8}{*}{Plants}                 & Press Space to use the shovel                                                      & 100\%                                & X                                                                        &                                                                \\
                                        & Use the shovel on grass to plant plants                                            & 100\%                                & X                                                                        &                                                                \\
                                        & Plants kill zombies by shooting pellets                                            & 76\%                                 &                              X                                            &                                                                \\
                                        & When plants get hit with axes, both are destroyed                                  & 53\%                                 & X                                                                        &                                                                \\
                                        & Protect the villagers from zombies for some time                                   & 35\%                                 & X                                                                        & X                                                              \\
                                        & Add plants to different areas to get good coverage                                 & 29\%                                 &                                                                          &                                                                \\
                                        & Axes don't affect player                                                           & 6\%                                  &                                                                          &                                                                \\ \cline{2-5} 
                                        & \textbf{Match Rate}                                                                & \textbf{-}                           & \multicolumn{1}{c|}{\textbf{81.8\%}}                                     & \multicolumn{1}{c|}{\textbf{11.9\%}}                           \\ \hline
\multirow{13}{*}{RealPortals}           & Press space to shoot a missile                                                     & 72\%                                 & X                                                                        &                                                                \\
                                        & If the missile collides with a wall, it turns into a portal                        & 72\%                                 & X                                                                        &                                                                \\
                                        & If a potion collides with water, the water is turned into ground                   & 72\%                                 & X                                                                        &                                                                \\
                                        & Unlock the door with the key                                                       & 68\%                                 & X                                                                        &                                                                \\
                                        & Collide with the goal to capture it                                                & 52\%                                 & X                                                                        & X                                                              \\
                                        & Collide with the key to pick it up                                                 & 48\%                                 & X                                                                        &                                                                \\
                                        & Pick up different wands to toggle between portal types                             & 44\%                                 & X                                                                        &                                                                \\
                                        & Teleport from the portal entrance to the portal exit                               & 44\%                                 & X                                                                        &                                                                \\
                                        & Collide with a potion to push it                                                   & 40\%                                 & X                                                                        &                                                                \\
                                        & Avoid dying by colliding with water or portal entrance with no exit                & 32\%                                 &                                                                          &                                                                \\
                                        & If a potion collides with the portal entrance, it is teleported to the portal exit & 16\%                                 & X                                                                        &                                                                \\
                                        & You can't go through the portal exit                                               & 0\%                                  & X                                                                        & \multicolumn{1}{l|}{}                                          \\ \cline{2-5} 
                                        & \textbf{Match Rate}                                                                & \textbf{-}                           & \textbf{94.3\%}                                                          & \textbf{9.3\%}                                                 \\ \hline
\end{tabular}
    }
    \caption{The \emph{Percentage} column designates the percentage of each mechanic being mentioned by humans in the user study. The X's in the \emph{Method} columns designate that the mechanic was included in the critical mechanic list for that method. The \emph{Match Rate} defines how closely this method agreed with human-identified critical mechanics. For all games, player movement (up-down-left-right) is an implied critical mechanic.}
    \label{tab:userstudyResults}
\end{table*}

Table~\ref{tab:userstudyResults} displays the results of both evaluations. In each game, for every critical mechanic that each discovery method identified, we record the percentage of users who believed the mechanic is important. We also include all other mechanics that participants thought are important but the discovery method does not. The ``Mechanic'' column contains the aggregated and summarized responses of the user study participants. Because the prompt was free-text, the exact wording of different game mechanics varied, but we attempted to approximate these into the mechanics of the game as they are written in a game's VGDL file. The ``Percentage'' column shows what percentage of the participants wrote down some form of this mechanic. For each of the games, we calculated each technique's match rate by summing the human-identified percentage value of the critical mechanics discovered by a method. That sum is then normalized over the summation of all percentages. The match rate therefore gives higher weight to the mechanics that more humans identified to be important. These values can be seen at the bottom of each game's section on Table \ref{tab:userstudyResults}.

The new playtrace method either is equivalent to or vastly improves over the baseline for every game when it comes to matching human opinion. Mechanics identified by the playtrace critical discovery method have the highest percentages of being mentioned by participants in all games except Solarfox. In Solarfox, a slightly higher number of people think that avoiding flames is more important than collecting the gems. We postulate that the constant movement of the player (the player can only change directions, not speed) and the large collision areas of the flames caused some users to focus more on flame avoidance than collecting gems. Humans not only identify important mechanics for winning but also ones to avoid losing. For example, in Zelda, ``Avoid dying by colliding with enemies'' is identified by 60\% of participants. Other participants note subgoals that usually reflect a better playing strategy, such as ``Add plants to different areas to get good coverage.'' The last mechanic type identified by participants pertains to scoring higher. In Zelda, the ``kill enemies with sword'' mechanic appears 76\% of time, and in Plants, the ``Plants kill zombies by shooting pellets'' mechanic also appears 76\% of time. Interestingly, the playtrace method does not classify this as a critical mechanic, instead opting to include plants getting hit with axes instead. We believe this is because plants shoot pellets independently of player actions, so by default planting more plants would result in more pellets and thus a higher chance of winning. Thus, this can be condensed down into just ``plant more plants.'' However, axes have a direct negative affect on plants and therefore impact a player's chance of winning. The algorithm found this shorter interaction path (``create plants'' - ``axes destroy them'') to be a simpler choice than including pellet interactions (``create plants'' - ``plants create pellets'' - ``pellets destroy zombies'').

One system-identified mechanic in Portals, ``You can't go through the portal exit,'' was never mentioned by any of the participants. We hypothesize there may be several reasons for this, one being that the mechanic seems very trivial to humans. It occurs in the playtraces because of the way the game is implemented in VGDL: after teleporting from entrance to exit, the game forces the player to step away from the exit. Participants who beat the game may not have thought it important enough to mention, and players who were unable to beat the game might have never realized that the portals were different types and colors.

\subsection{Agent Performance Study}\label{sec:validation-agents}
In this evaluation, we compare the performance of an MCTS agent with no mechanic information (vanilla) against MCTS agents augmented with the critical mechanics for Solarfox, Zelda, Plants, and RealPortals discovered using the \emph{AtDelfi} method~\cite{green2018atdelfi} and the new playtrace method presented in this paper. The vanilla MCTS agent is a clone of the MCTS agent that comes with the GVG-AI framework and used for benchmarking in other GVG-AI projects. Agents given critical mechanic information have an identical configuration to the vanilla agent, with the sole exception being the Reward Calculation, which is replaced with the process explained in Section \ref{sec:overview-mechanic} instead of game score. Finally, a second benchmarking agent is given all the mechanics for each game, also being rewarded each time any of these mechanics are triggered. The C value for all agents in the UCT equation was fixed to $0.125$. Regardless of mechanic information given, each agent is given $5$ unique levels to play for each game, $3$ of these levels being identical to the user study levels and $2$ being unique to this evaluation. Each level is played $20$ times, for a total of $100$ playthroughs per game. An agent is permitted to build a search tree of up to 5000 nodes before deciding its next action every turn. An agent is permitted a maximum rollout of 50 moves for each node expansion. All experiments took place on Intel Xeon E5-2690v4 2.6GHz CPU processor within a Java Virtual Machine limited to 8GB of memory. An experiment was allowed to be a maximum of 48 hours long; however, none reached this limit. 

\begin{figure*}[t]
    \centering
    \begin{subfigure}[t]{.48\linewidth}
        \centering
        \includegraphics[width=\linewidth]{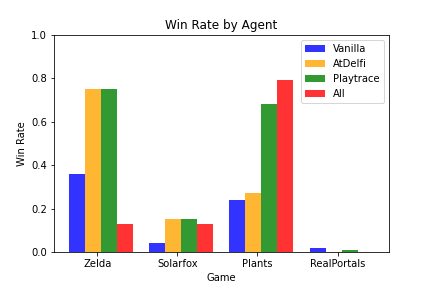}
        \captionsetup{width=.95\linewidth}
        \caption{The win rates of agents on all four games.}
        \label{fig:win-rate}
    \end{subfigure}
    \begin{subfigure}[t]{.48\linewidth}
        \centering
        \includegraphics[width=\linewidth]{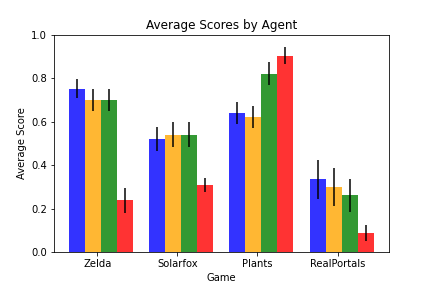}
        \captionsetup{width=.95\linewidth}
        \caption{The mean normalized scores of agents on all four games.}
        \label{fig:scores}
    \end{subfigure}
    \caption{Comparing the performance between the different agents.}
    \label{fig:comparison}
\end{figure*}

Figure~\ref{fig:win-rate} displays a comparison of win rates between the agents, and Figure~\ref{fig:scores} displays average normalized scores with a $95\%$ confidence interval. Scores are normalized by level using the maximum and minimum obtainable scores for that level and then averaged together. Zelda and Solarfox both have fixed maximum and minimum scores for all levels. Because the maximum score value in Plants is based on randomness, we instead score agent performance by their survival time. RealPortals does not have an upper bound on score due to the nature of its game mechanics, so we clamp scoring to the minimum optimal score needed to solve each level.

From Figure~\ref{fig:comparison}, it can be seen that the playtrace method was able to achieve better performance than the vanilla MCTS on all games, and better performance than the \emph{AtDelfi} method on Plants. The \emph{AtDelfi} method appears to have a higher average score on RealPortals. However, due to the confidence interval for both of the augmented agents, we can assume that the score difference between the two is most likely the result of random noise. The low win rates in RealPortals may be a response to the complexity of the game. Because of this complexity, achieving a higher score is a mixed signal: it could mean the agent is closer to winning, but it could also mean the agent is simply abusing the game rule of repetitively going through a portal to get more points. The \emph{All} mechanics agent seems to perform better on Plants, with comparable performance in Solarfox and the worst performance in Zelda.

\section{Discussion}
Based on the results from the user study and the agent experiments, we conclude that the new playtrace method is more successful than the \emph{AtDelfi} method at correctly identifying critical mechanics. This suggests that the method could be a crucial component in a tutorial generation system.

For Zelda, Solarfox, and Plants, the Playtrace agent results demonstrate significant win-rate improvement over the vanilla MCTS agent when critical mechanics are incorporated into the search algorithm. In particular, the Playtrace method outperformed \emph{AtDelfi} in Plants, due to the inclusion of some highly important mechanics about planting defenses. None of the methods help agents win RealPortals, suggesting there is room for improvement in how this information is incorporated into the agent. This is further supported by the inconsistant gameplay of the All agent, which is rewarded for any game mechanic being triggered. In a game like Plants, where there are $15$ mechanics in total, incorporating every mechanic seems to have a strong positive affect. But in Zelda, which contains nearly three times as many, this causes the opposite. We speculate this has something to do with how the MCTS agents are rewarded for mechanic triggers. For Plants, most mechanics are directly or indirectly caused by the player planting more plants. Any action, therefore that involves planting a plant is highly rewarded. In Zelda, however, a good portion of the mechanics involve enemies bumping into walls and each other. When every branch in the tree is rewarded for these stochastic occurrences, MCTS agents behave more like breadth first search agents. The AtDelfi and Playtrace methods overcome this by allowing the agent to focus on the game-winning mechanics only, and therefore can take advantage of MCTS' ``exploitation'' factor.

The new playtrace method demonstrates matched or significant improvement over the \emph{AtDelfi} method based on the match rates shown in Table \ref{tab:userstudyResults}. Interestingly enough, although the playtrace method has its highest match rate with RealPortals, an agent augmented with those mechanics only manages to win the game 1\% of the time. This situation proves that a two-step evaluation procedure provides a deeper understanding than either being a stand-alone process. In the context of tutorial generation for humans, a method which helps an AI achieve a stable win rate yet fails to address many of the human-identified critical mechanics cannot be considered very successful.

Humans identify important mechanics not present in the playtrace method's critical mechanic set, like the fact that the player can kill enemies in Zelda, that one should avoid flames in Solarfox, or that peas kill zombies in Plants. We can attribute this to the way our system searches for the critical path. The goal of the system is to find a \textit{least cost path} using mechanics that result in a winning state, thus it \emph{does not} search for a result that \emph{avoids} a losing state. As a result, it will not actively include mechanics that may be important to players in order to avoid dying or losing the game (such as avoiding flames in Solarfox). We had expected the playtrace method to include mechanics like slaying monsters in Zelda or that peas slay zombies in Plants. Due to the way playtrace information is inserted (only one playtrace with the minimum amount of noise is used), the critical mechanics may change, depending on what happened this particular playtrace and when mechanics were triggered in relation to each other.



Our method is focused on mechanics being triggered during play, but what it admittedly fails to capture are any mechanics one would \emph{not} want to trigger to win. Solarfox best exemplifies this, where running into walls or flames would cause a loss, i.e. make it impossible to win. Players believed this to be important to mention in Table \ref{tab:userstudyResults}. We limited the scope of this paper to include only these ``positive'' mechanics, as we believe that discovering ``negative'' mechanics is a non-trivial problem by itself. Hence, our definition of critical mechanic limits itself to mechanics that must be triggered to win. We believe this problem is a research question by itself, and plan to improve our approach to include it in future work.

There is an interesting discussion point to be had in regards to a game like Realportals. Even though agent performance is higher on a scoring basis, neither augmented agent can reliably win levels, and the way that it is gaining points (going back and forth between portals repetitively) can hardly be considered a successful strategy for a human being. Despite this, users strongly concurred with the playtrace method's mechanics, suggesting that for this game (and perhaps others similar to it in complexity) the agent will have to be more intelligently augmented with mechanics. 



\section{Conclusion}
In this work, we present new method for automatically discovering critical mechanics from games using playtraces. We perform a two-step procedure for evaluating all future critical mechanic discovery methods. First, we use human intuition as one evaluator for critical mechanic discovery. The new playtrace method is compared to the \emph{AtDelfi} method using a match rate to human-identified mechanics. In Solarfox and Zelda, the methods identify the same critical mechanics, so there was no change in match rate. However, in both Plants and RealPortals, the playtrace method has a much higher match rate.
We also use these mechanics to augment MCTS agents to observe how game-play performance improves. In two of the tested games (Zelda and Solarfox), the playtrace method agent shows matched performance to the \emph{AtDelfi} method agent. In Plants, the playtrace method agent shows massive improvement over the \emph{AtDelfi} method agent. In RealPortals, although both methods obtain higher average scores than the vanilla agent, neither the \emph{AtDelfi} nor the playtrace method agent is able to win the game a significant amount of times.

This work can be used to further research in mechanic discovery and mechanic usage in games and game applications. By using past playtraces and game time as units of measure, our system is able to identify mechanics and augment MCTS agents with them, improving agent performance. These mechanics might be able to augment agents in other ways too, like using them as intermediate rewards during training to help reinforcement learning agent generalize better. We believe this research could be a foundation for an intelligent debugging process for game developers, allowing them to adjust a game's rules/levels in response to the playtrace of an agent augmented with the mechanics of the game. This work is compatible with the idea of game state compression~\cite{cook2019hyperstate}, in the sense that mechanics which could be defined as causing ``irreversible states''. Hyperstate analysis might give insight into which mechanics should be considered ``critical'' or vice versa.

Our critical mechanic discovery method is primarily meant to be used within tutorial generation, such as the \emph{AtDelfi} system~\cite{green2018atdelfi}, to automatically construct tutorials that teach \textit{humans} how to play games. Prior research~\cite{green2018generating,khalifa2019intentional} demonstrates that mechanics can be used to generate levels, allowing the mechanics found here to be used in that process. In addition to the arcade games shown here, our system could be extended in future work to incorporate more complex games. Our system can capture macro actions in these larger goal-oriented games (and in ones where players define their own goals), which can then be used to extract the critical mechanics as demonstrated in this paper. Although the playtrace method presented in this paper shows improvement over the \emph{AtDelfi} method, we postulate that there are other, better methods to be created. Furthermore, we plan on improving existing tutorial generation systems by automatically generating instructions and levels that teach game mechanics using this approach. 

\begin{acks}
Michael Cerny Green acknowledges the financial support of the GAANN program. Ahmed Khalifa acknowledges the financial support from NSF grant (Award number 1717324 - ``RI: Small: General Intelligence through Algorithm Invention and Selection.''). Tiago Machado acknowledges the finacial support from CNPq - Conselho Nacional de Desenvolvimento Cient\'{i}fico e Tecnol\'{o}gico under the Science without Borders scholarship 202859/2015-0.
\end{acks}

\bibliographystyle{ACM-Reference-Format}
\bibliography{sample-base}

\end{document}